\documentclass[letterpaper, 10 pt, conference]{ieeeconf}  
\usepackage{etoolbox}
\usepackage{subcaption}
\usepackage{amsmath,amsfonts}
\usepackage{algorithm}
\usepackage{array}
\usepackage{textcomp}
\usepackage{stfloats}
\usepackage{url}
\usepackage{verbatim}
\usepackage{graphicx}
\usepackage{cite}

\usepackage[table,xcdraw]{xcolor}
\usepackage{algpseudocode}
\usepackage{multirow}
\usepackage{cite}
\usepackage{amsmath,amssymb,amsfonts}
\usepackage{graphicx}
\usepackage{textcomp}
\usepackage{xcolor}
\usepackage{color}
\usepackage{algorithm}
\usepackage{algpseudocode}
\usepackage{authblk}
\usepackage{hyperref}
\usepackage{comment}
\usepackage{subcaption} 

\usepackage{fancyhdr}
\fancypagestyle{withfooter}{
  
  \fancyhead[L]{}
  \fancyhead[R]{}
  \fancyfoot[C]{\footnotesize Presented at the 2025 IEEE ICRA Workshop on Field Robotics}
}

\definecolor{blue}{rgb}{0, 0, 1}

\definecolor{yellow}{rgb}{0.7, 0.7, 0}

\IEEEoverridecommandlockouts                              

\begin{document}

\title{\LARGE \bf
Learning Rock Pushability on Rough Planetary Terrain
}

\author{Tuba Girgin*, Emre Girgin*, Cagri Kilic
\thanks{This study was supported by 2024-2025 COE-BCAAS Research
Stimulus Program. }

\thanks{*Equal contribution. }

\thanks{The authors are with the Department of Aerospace Engineering at Embry-Riddle Aeronautical University in Daytona Beach, Florida, USA.
        {\tt\small cibukgig@my.erau.edu, girgine@my.erau.edu, kilicc@erau.edu}}%
}

\maketitle
\thispagestyle{withfooter}
\pagestyle{withfooter}

\begin{abstract}
In the context of mobile navigation in unstructured environments, the predominant approach entails the avoidance of obstacles. The prevailing path planning algorithms are contingent upon deviating from the intended path for an indefinite duration and returning to the closest point on the route after the obstacle is left behind spatially. However, avoiding an obstacle on a path that will be used repeatedly by multiple agents can hinder long-term efficiency and lead to a lasting reliance on an active path planning system. In this study, we propose an alternative approach to mobile navigation in unstructured environments by leveraging the manipulation capabilities of a robotic manipulator mounted on top of a mobile robot. Our proposed framework integrates exteroceptive and proprioceptive feedback to assess the push affordance of obstacles, facilitating their repositioning rather than avoidance. While our preliminary visual estimation takes into account the characteristics of both the obstacle and the surface it relies on, the push affordance estimation module exploits the force feedback obtained by interacting with the obstacle via a robotic manipulator as the guidance signal. The objective of our navigation approach is to enhance the efficiency of routes utilized by multiple agents over extended periods by reducing the overall time spent by a fleet in environments where autonomous infrastructure development is imperative, such as lunar or Martian surfaces.

\end{abstract}

\section{INTRODUCTION}

In autonomous mobile robot navigation, obstacle avoidance is critical and extensively
studied in both dynamic and static contexts \cite{Rafai2022, Guo2022, Choi2021}. However, avoiding obstacles often increases travel time or it may not even be possible to cross within the given obstacle avoidance policy, which can be problematic in scenarios requiring repeated traversal, such as sample return missions on planetary surfaces \cite{Morota2020, Lauretta2022}, post-disaster search and rescue operations \cite{Cruz2021}, or navigation in hazardous environments like nuclear power plants \cite{Groves2021}.  In such cases, a more efficient option could be to remove obstacles that are easily displaceable. Manipulating the environment manually, such as astronauts performing tasks on the lunar surface, is time-consuming, limits their ability to focus on primary missions, and can be dangerous in certain areas. Therefore, autonomous robotic manipulation is necessary. However, this introduces new challenges, such as autonomously determining whether an obstacle is worth relocating and whether it can be relocated, which is non-trivial due to unknown physical constraints. These constraints include obstacle characteristics \cite{Di2013}, surface properties such as slope and slipperiness \cite{Teji2023}, and the medium through which the obstacle is manipulated \cite{Suomalainen2022}.

The manipulation of objects in office or tabletop environments has been extensively studied \cite{ozdamar_pushing, heins_push, yang_sim}. However, the interaction between the objects being manipulated and the surfaces they rest on remains underexplored. Most studies on object manipulation focus solely on the properties of the objects \cite{ahmetoglu_deepsym, sahin_2007} without considering the influence of diverse surface properties. Moreover, research that does take surface properties into account is largely limited to surfaces with known characteristics \cite{Lloyd_2021}.

\begin{figure}[t]
    \centering
    \includegraphics[ width=0.5\textwidth]{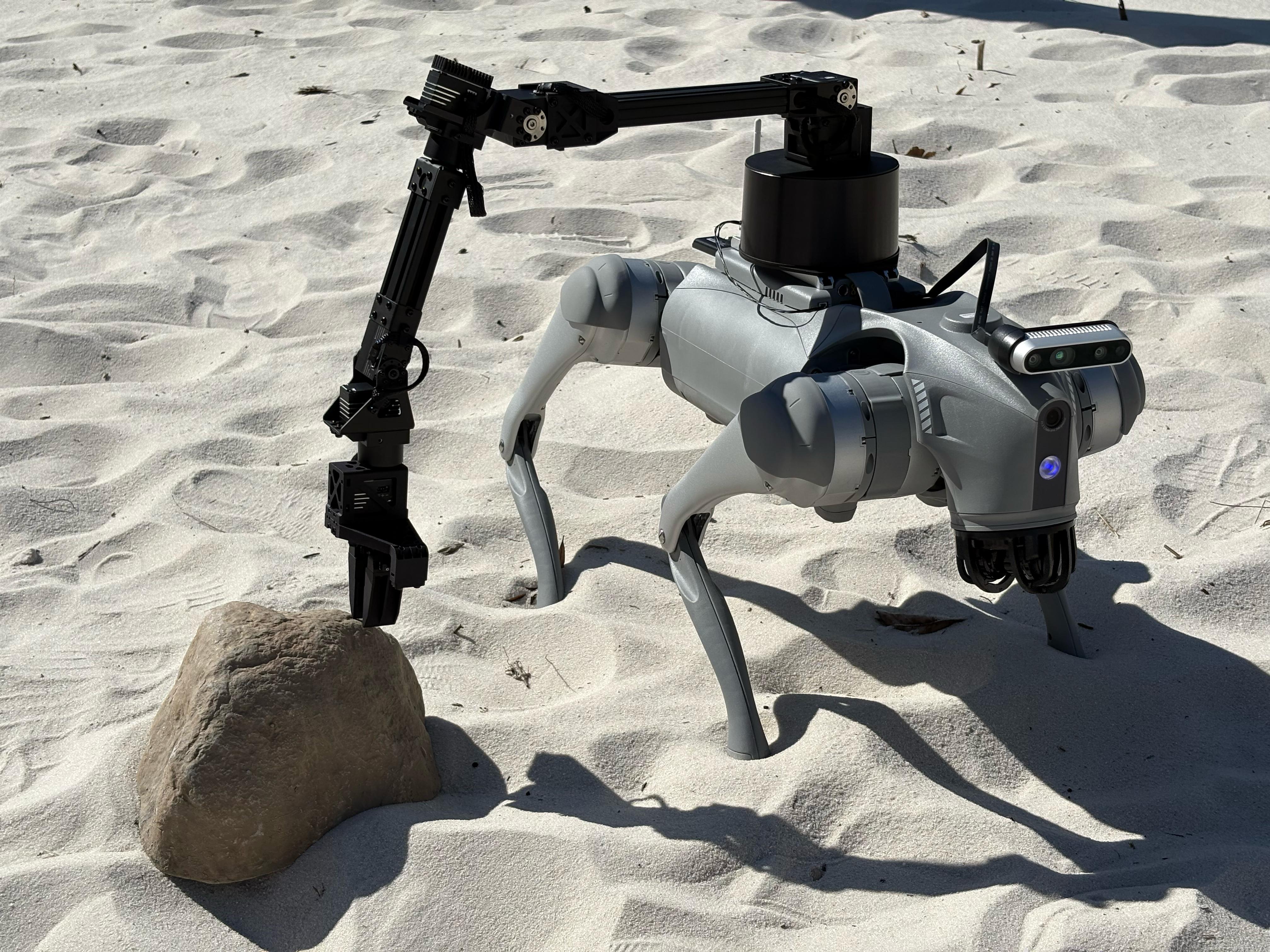}
    \caption{A depiction of rock pushing in rough terrains is shown. The goal is to clear pushable rocks to enhance the effectiveness of repeated navigation for various robots.}
    \label{fig:robot}
\end{figure}

We propose a probabilistic affordance learning framework to determine whether an
obstacle in the robot's path can be removed using exteroceptive information and proprioceptive feedback. In robotics, the affordance of an object refers to the action possibilities it offers to an agent \cite{Jamone2016}. The action possibilities an obstacle offer depends on its physical properties, the surface it relies on, and, in our case, the characteristics of the pushing system. While visual cues like size and material suggest potential manipulation, the physical interaction between an object and variable surfaces, especially on planetary terrains with spatially varying slopes and soil characteristics \cite{Arvidson2016, Mittlefehldt2021}, often necessitates active exploration \cite{Seminara2019}. Our framework addresses this issue by utilizing a robotic arm to interact with obstacles, combining visual data of the obstacle and the surface it rests on, with force feedback data to estimate whether the interested obstacle can be relocated.

In this study, we designed a framework utilizing Bayesian Linear Regressor in which obstacle and surface features are extracted from the scene point cloud using cosine similarity-based segmentation and clustering with DBSCAN. The proprioceptive force feedback is predicted for the obstacles being actively explored by the pushing action. We designed a simulation environment where the terrain consists of hills and pits generated by Perlin noise, accompanied by a variety of rocks. In the real-world setup, we prepared environments featuring different surface slopes and obstacle sizes. We presented the resulting dependency graph, which shows the relationship between the observed scene, proprioceptive feedback, and push affordance. We collected an active exploration dataset that includes point cloud and proprioceptive force feedback data, gathered from both simulation environments and real-world setups. The code and dataset will be published at the following GitHub link: https://github.com/srge-erau/rock-push-affordance.

\section{RELATED WORK}

Planetary exploration and navigation have been widely studied in the literature. In the field, visual odometry \cite{bahraini_autonomous, wudenka_toward, Mahlknecht_event, zhou_innovative, lu_landmark}, visual SLAM \cite{hong_visual, romero_evaluation}, and semantic segmentation \cite{kuang_rock, kuang_semantic, kuang_sky} are extensively studied, primarily focusing on camera images. When relying solely on visual data is insufficient for planetary exploration tasks, proprioceptive sensor feedback is used to enhance performance. Sinkage and slippage detection have been widely studied, either using only proprioceptive sensor feedback \cite{gonzales_slippage, coloma_enhancing, kilic_pro, kilic_slip, kilic_improved} or by combining it with visual data \cite{feng_learning} to enhance navigation on planetary surfaces. Terrain estimation \cite{Dimastrogiovanni_terrain} is also another research area that is widely studied in the literature that utilizes proprioceptive sensing.

 One common aspect of the aforementioned areas of study is that they involve passive exploration of the environment, without interacting with surrounding obstacles to further analyze the characteristics of the observed scene. On the other hand, object interaction and manipulation studies focuses on object grasping \cite{zhang_graspability}, non-prehensile pushing \cite{heins_push, yang_sim}, and learning object features with these actions \cite{ahmetoglu_deepsym, girgin_multi}.

 King et al. \cite{king_nonprehensile} focused on rearrangement planning of daily life objects in a table top setup. Tekden et al. \cite{tekden_object} studied nonprehensile pushing of object clusters to learn the relation graph between the objects on a flat surface. Ozdamar et al. \cite{ozdamar_pushing} utilized a robotic arm on a wheeled base to goal-based pushing of big objects like boxes and cylinders. These studies focus on object manipulation on flat surfaces such as tables and floors, while the interaction between planetary terrains and obstacles, such as rocks of various sizes and shapes, necessitates an analysis of the surface on which the object rests.

Our research fills this gap by integrating visual data and proprioceptive sensing to estimate pushability of obstacles in unknown environments, particularly on complex planetary surfaces where traditional navigation methods may prove inadequate.



\begin{figure*}[t]
    \centering
    \includegraphics[trim=0cm 2.8cm 0cm 1.3cm, clip, width=15cm]{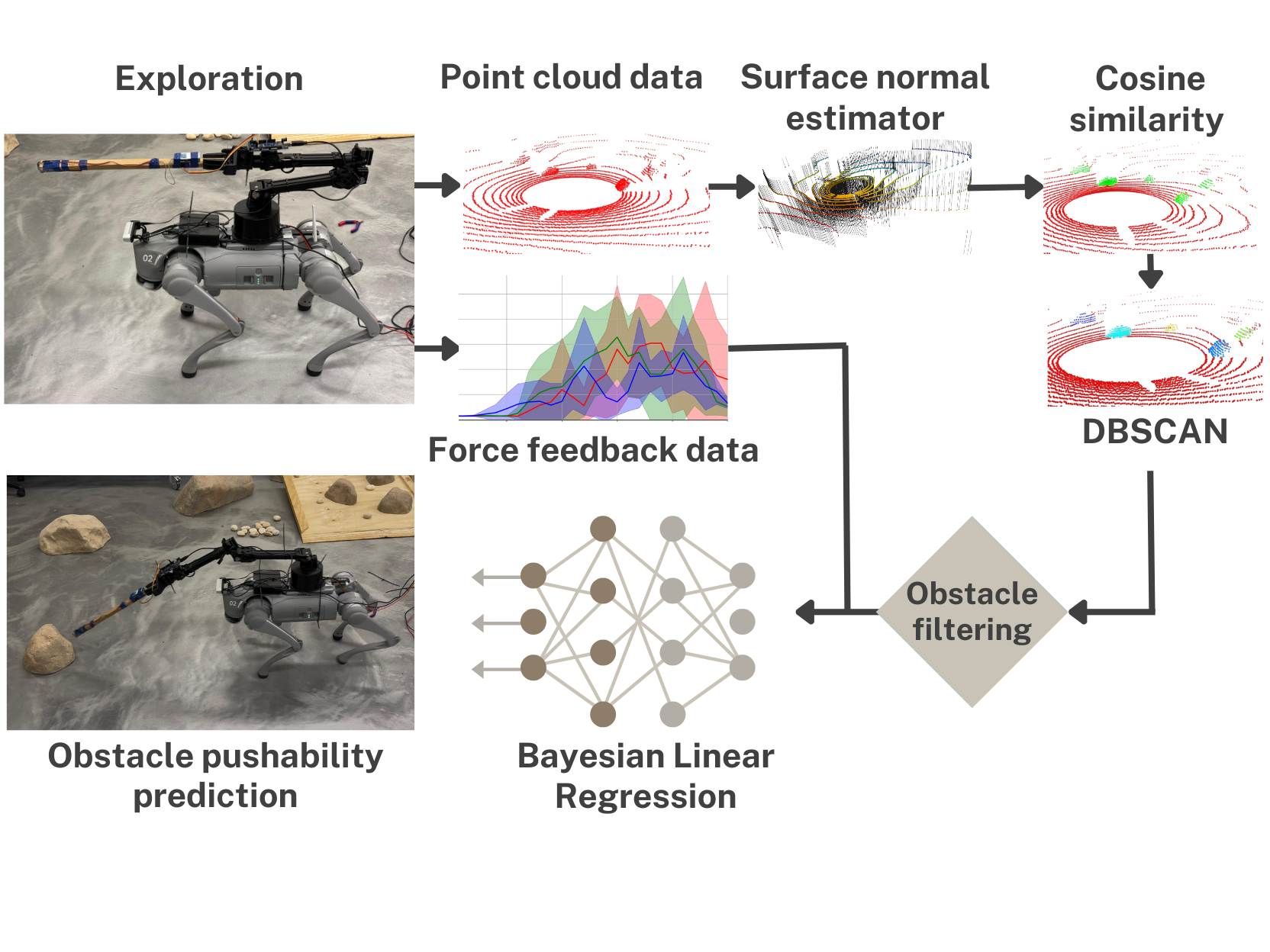}
    \caption{The architecture of our proposed pushability learning probabilistic model, which incorporates both visual and proprioceptive data, is demonstrated. The robot actively explores its environment, collecting point cloud data and proprioceptive force feedback while pushing the target rock. Surface normals are estimated from the point cloud data, and rocks are detected using the cosine similarity method. An unsupervised DBSCAN model is employed to cluster the rocks. Obstacle filtering is applied to eliminate rocks that are either too large or too small. Using the extracted features from the scene, a Bayesian linear regression model is fitted to predict the maximum magnitude sensed during the pushing action, which has an inverse relationship with the pushability of the rock. }
    \label{fig:flow}
\end{figure*}

\section{METHOD}

\subsection{Problem Definition} 

At a given time, the scene is observed by the  robot as a point cloud, which can be represented as a set of points:
 \(
\mathcal{P} = \{ p_1, p_2, \dots, p_n \}, \quad p_i \in \mathbb{R}^3.
\) where \( n \) is an arbitrary positive integer. An obstacle \( \mathcal{O}_j \) is a subset of the point cloud containing an arbitrary number of points:
\(
\mathcal{O}_j \subseteq \mathcal{P}, \quad j = 1, 2, \dots, m,
\) where \( m \) is the arbitrary number of obstacles. The obstacles do not share any points, meaning they are mutually disjoint:
\(
\mathcal{O}_i \cap \mathcal{O}_j = \emptyset, \quad \forall i \neq j.
\) The remaining points in the point cloud, which do not belong to any obstacle, are defined as the ground:
\(
\mathcal{G} = \mathcal{P} \setminus \bigcup_{j=1}^{m} \mathcal{O}_j.
\)

The 3D bounding box \( \mathcal{B}_j \) for the \( j \)-th object can be defined as the axis-aligned bounding box (AABB) enclosing the object:
\(
\mathcal{B}_j = [s \cdot x_{min}^j, s \cdot x_{max}^j] \times [s \cdot y_{min}^j, s \cdot y_{max}^j] \times [s \cdot z_{min}^j, s \cdot z_{max}^j],
\)
where \( s > 1 \) is the scaling factor, \( (x_{min}^j, y_{min}^j, z_{min}^j) \) and \( (x_{max}^j, y_{max}^j, z_{max}^j) \) are the coordinates of the minimum and maximum corners of the initial bounding box, respectively. 
The surface \( \mathcal{S}_j \) on which the object \( \mathcal{O}_j \) rests is defined by the ground points remaining within the bounding box \( \mathcal{B}_j \) :
\(
\mathcal{S}_j = \{ p \in \mathcal{G} \mid x_{min}^j \leq p_x \leq x_{max}^j, \; y_{min}^j \leq p_y \leq y_{max}^j, \; z_{min}^j \leq p_z\},
\) where \( p = (p_x, p_y, p_z) \) .

Surface normal set of surface \(\mathcal{S}_j\) is \( \mathcal{N}_j = \{n_1, n_2, n_3, \dots n_k \} \), where \(k = \lvert S_j \rvert \) and \(n_i\)'s are normalized unit vectors. Size of an obstacle is \( \mathcal{V}_j = \Delta x \cdot \Delta y \cdot \Delta z \), which is equal to the volume of the 3D bounding box when scale factor \(s = 1\). To represent the shape of a rock \(\mathcal{E}_j \in \mathcal{R}\), we adopt the approach where we fit the rock into ellipsoids and score based on ellipse fitting error \cite{Di2013}.


The robot interacts with the obstacle of interest applying a nonprehensile push action. In this study, we assume our robot is capable of performing a predesigned push action with finite force. The robot's physical capabilities \(\delta\) include its size and maximum force it can apply. The feedback force applied to the robotic arm by the obstacle being pushed \( \mathcal{O}_j \) is captured by the proprioceptive sensor attached to the arm. The resulting force signal  \( \mathbf{F^j} = [f_1^j, f_2^j, \ldots, f_{t}^j], \quad f_i^j \in \mathbb{R}^3 \) consists of 3D force values where \( t \) is a predetermined constant. The result of a push action is defined as the maximum magnitude in the force signal: \( f_{max}^j = \max_{1 \leq i \leq t} \|\mathbf{f}_i^j\|_2, \) where \( \|\mathbf{f}_i^j\|_2 \) denotes the Euclidean norm of the \( i \)-th 3D force value in the force signal \( \mathbf{F^j} \). The pushability affordance of an obstacle \(\mathcal{O}_j \) is defined as the probability of it moving by predicting the force feedback conditioned on the size \(\mathcal{V}_j\) and shape \(\mathcal{E}_j\) of the object and the normal surface \(\mathcal{N}_j\) such that \(\mathcal{A}_j = ( f_{max}^j | \mathcal{V}_j, \mathcal{E}_j, \mathcal{N}_j)\).

In this study, our goal is to predict the pushability affordances, \(\mathcal{A} \), of obstacles, \(\mathcal{O} \), on the terrain, \(\mathcal{G} \). This allows the robot to decide which objects can be pushed.

\subsection{Methodology} \label{sec:methodology}

Our architecture consists of two main parts where the first conducts a preliminary analysis to filter an obstacle \(\mathcal{O}_j\) based on visual features extracted from \(\mathcal{P}\) and the second stage preditcs the pushability of it by interacting physically and exploiting proprioceptive feedback. Figure \ref{fig:flow}. summarizes the pipeline starting from the point cloud of the scene and the force feedback received by interaction to the Bayesian inference. The visual examination section consists of surface normal estimation, point cloud segmentation, obstacle characterization, and preliminary prediction. The affordance learning section ignores the filtered out obstacles which are either too small or too big to be interacted safely and predicts the pushability affordances for each. The affordance prediction utilize surface normal and the obstacle characteristics as an input to bayesian learner and estimates the probability of required force.  

\subsubsection{Visual Examination}

The visual feature extraction and preliminary prediction section $ VPP(\delta, \mathcal{V}_j, \mathcal{E}_j, \mathcal{N}_j) = \{S, P, O\}^m\) use features extracted from pointcloud \(P\) and robot's capabilities $\delta$ to classify obstacles according to visual characteristics.  

In order to distinguish points belonging to the ground $\mathcal{G}$ and obstacle \(\mathcal{O}_j\) a surface normal based segmentation approach is utilized. Once the surface normals \( \mathcal{N}_{scene} \in \mathcal{R}^{n \times 3} \) are calculated for each point in \(\mathcal{P}\), each point is segmented whether it belongs to the \(\mathcal{G}\) or an obstacle \(\mathcal{O}_j\) based on its normal's cosine similarity to the median surface normal of the scene. Those that exceed a certain threshold \(T_{cs}\) are considered ground points.  

Rigorously, the surface normal \( n_i \) of the point \(p_i \in \mathcal{P}\) is the eigenvector corresponding to the minimum eigenvalue of covariance matrix \(\mathcal{C}_i\) formed by the \(k\) nearest neighbors of \(p_i\) within the radius \(r\). Thus, the neighbor point set of $p_i$ is denoted as \(\mathcal{H}_i^{(k,r)} \).

The covariance matrix \(\mathcal{C}_i\) of neighborhood \(\mathcal{H}_i^{(k,r)} \) is \(\mathcal{C}_i = \frac{1}{\lvert \mathcal{H}_i \rvert } \sum_{p_q} (p_q - c_i) \cdot (p_q - c_i)^T \), where \(c_i = \frac{1}{\lvert \mathcal{H}_i \rvert} \sum_{p_q} p_q\) is the centroid of neighborhood and \(p_q \in \mathcal{H}_i^{(k,r)}\). Then \(n_i\) is the normalized eigenvector corresponding to the smallest eigenvalue of \(\mathcal{C}_i\).

We assume the median surface normal \(n_{med}\) of the point cloud represents the ground normal of the scene. Since the surface normals may point inward or outward, we use cosine similarity between each normal vector and the \(n_{med}\) to eliminate the sign inconsistency. Therefore ground points \(\mathcal{G}\) is equal to \(\{ p_i \in \mathcal{P} | g(n_{med}, n_i) \geq T_{cs} \} \), where \(g\) is the cosine similarity function.


To distinguish obstacles from one another, DBSCAN algorithm (\( \text{DBSCAN}(\mathcal{P} \setminus \mathcal{G}; \epsilon, k) = \{l_1, l_2 \dots l_m\}^u \), where $k$ and $\epsilon$ are hyperparameters ) is utilized on the points which do not belong the ground.  We assume the remaining points belonging to the different obstacles will be spatially distinct and discontinuous from each other after the ground filtering. Therefore, each cluster contains only the points belonging to the same obstacle. Therefore, obstacle \(O_j\) is \(\{p_i | d(p_i; \epsilon, k) = l_j  \} \), where \( l_j \) is cluster number of obstacle \(\mathcal{O}_j\). Note that, the points can not be clustered due to lack of enough points around to form a cluster is labeled as $-1$ and considered as a ground point. 


Once the obstacle \(O_j\) is formed, the size and shape features are extracted. Whereas size \(V_j\) is equal to the volume of the 3D bounding box when scale factor \(s = 1\), the shape  \( \mathcal{E}_j \) is calculated by the fitting a 3D ellipsoid to the obstacle \(O_j\) as an extension of \cite{Di2013}'s approach from 2D to 3D. The shape of an obstacle is defined as the root mean residual error of the fitted ellipsoid. We evaluate ellipse fitting as a least-square problem and we fit the points belonging to the same obstacle to a ellipsoid by minimizing algebraic distance to solve ellipsoid coefficients. Therefore angularity is derived from the root mean residual error \(e_j\) of the ellipsoid fit. The errors close to zero means near match an idealized ellipsoid and the error increases as the obstacle has more sharp edges or corners.

\begin{figure}[h]
    \centering
    \begin{subfigure}[t]{0.24\textwidth}
        \centering
        \includegraphics[height=1.2in,width=\textwidth,keepaspectratio]{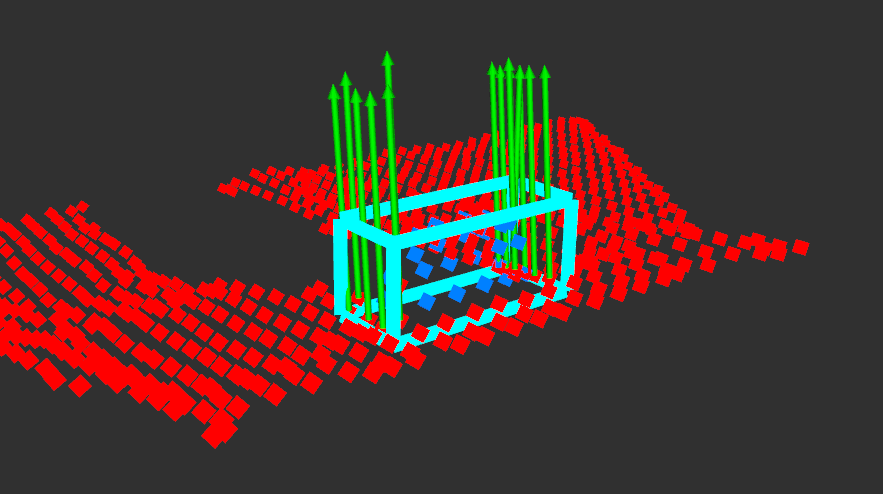}
        \caption{Downhill Surface}
        \label{fig:downhill_normal}
    \end{subfigure}%
    \hfill
    \begin{subfigure}[t]{0.24\textwidth}
        \centering
        \includegraphics[height=1.2in,width=\textwidth,keepaspectratio]{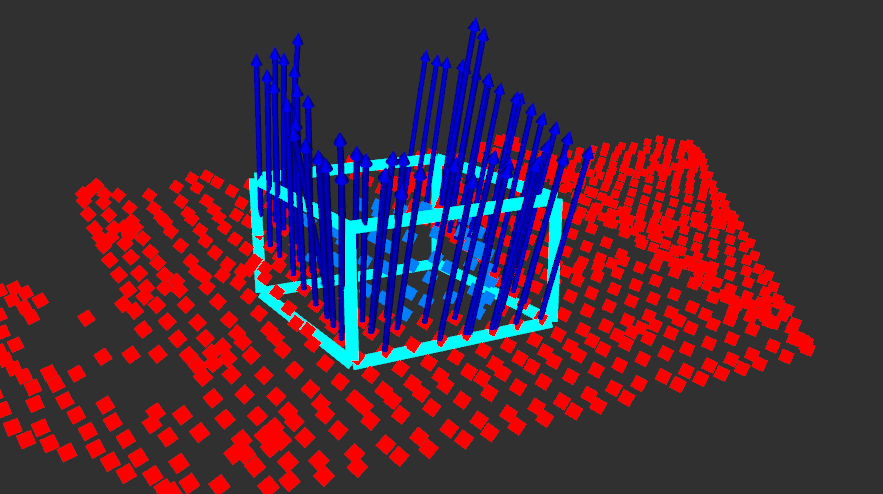}
        \caption{Uphill Surface}
        \label{fig:uphill_normal}
    \end{subfigure}
    \caption{VPP can determine the normal of surface the obstacle relies on, in real-world environment. Green and blue arrows represent the normals having $\theta < \pi/2$ and $\theta > \pi/2$, respectively.}
    \label{fig:surface_normals}
\end{figure}

Surface \(\mathcal{S}_j\) of obstacle \(\mathcal{O}_j\) is the set of ground points in the corresponding bounding box \(\mathcal{B}_j\) is scaled by \(s > 1\). \( \mathcal{S}_j = \{p_i | p_i \in \mathcal{G} \land p_i \in \mathcal{B}_j^s \land p_i \} \).  The surface normal \(\mathcal{N}_j\) is the corresponding surface normals of the points in \(\mathcal{S}_j \) such that \(\mathcal{N}_j = \{n_i | p_i \in \mathcal{S}_j\} \) (Fig \ref{fig:surface_normals}).



Finally, the visual estimation system processes the size \(\mathcal{V}_j\), shape \(\mathcal{E}_j\), and surface normals \(\mathcal{N}_j\) of obstacle \(\mathcal{O}_j\) to classify it into one of three categories:
\begin{itemize}
    \item S (Static): too large to be pushed,
    \item P (Pushable): likely to be pushed,
    \item O (Override): too small to be avoided.  
\end{itemize}

Since these classifications depend on the robot’s physical capabilities \(\delta\), such factors are also considered. The visual preliminary prediction function assigns a likelihood score \(g_j\) to obstacle \(\mathcal{O}_j\) and classifies it as:

\begin{equation} \label{eqn:filtering}
VPP(\mathcal{V}_j, \mathcal{E}_j, \mathcal{N}_j, \delta) =
   \begin{cases} 
      \text{S}, & g_j \leq T_{\text{low}} \\  
      \text{P}, & T_{\text{low}} < g_j < T_{\text{high}} \\  
      \text{O}, & g_j \geq T_{\text{high}}  
   \end{cases}
\end{equation}
where S denotes Static, P denotes Pushable, and O denotes Override. The likelihood score \(g_j\) is proportional to the capabilities of the robot and inverse proportional to size and shape. The surface normal helps or hurts the pushability depending on the relative position and orientation of the obstacle with respect to the robot. The \(p_i\) is also the displacement vector from the robot to the surface normal \(n_i\). The angle \(\theta\) between the displacement vector \(p_i\) and its normal \(n_i\) helps the pushability about $0$ and starts to hurt as it approaches to $\pi$. Therefore, the likelihood is designed as \(g_j = \frac{\delta}{\mathcal{V}_j \cdot \mathcal{E}_j}(1 - \frac{\theta}{\pi})\). For simplicity, we use the mean of the \(\mathcal{O}_j\) as displacement vector and \(\mathcal{N}_j\) as the surface normal, at this stage. 

\subsubsection{Probabilistic Affordance Learning}

Bayesian models in machine learning are known for their built-in capability of representing prior knowledge and probabilistic uncertainty quantification. While the prior distribution assumption aligns with our framework's preliminary visual prediction phase, we also utilize its ability to model prediction as a posterior distribution and measuring the uncertainty.

Therefore, the second section of our framework adopts a Bayesian regressor that predicts the affordance of an obstacle as a probability distribution given the visual preliminary prediction (VPP). The output of the VPP serve to update a prior belief into a posterior distribution over the Bayesian models's parameters. The prediction of the samples generate a full distribution which naturally quantifies uncertainty.

We condition the pushability affordance on the proprioceptive feedback produced while the robotic manipulation of the obstacle. We perform a push action towards the obstacle and measure the force feedback. Our expectation is to observe a proprioceptive pattern within the mobile and stationary effects. While stationary obstacles cause a feedback up to the limits of the robotic arm, the pulse generated by the push action towards the pushable obstacles reaches a peak until it overcomes the static friction and gradually decreases as the obstacle moves. The Bayesian model predicts the peak value of the force feedback and pushability affordance of the obstacle is determined if the prediction is within the robot's capabilities \(\delta\).

\begin{algorithm}[H]
\caption{Probabilistic Bayesian Affordance Learning}
\begin{algorithmic}[1]
\Procedure{BayesianAffordance}{TrainSet, TestSet, $\delta$ }
    \text{[\(R_i = (\mathcal{V}_i, \mathcal{E}_i, \mathcal{N}_i, g_i)\)]}

    \textbf{1-) Training Phase}
    \State \(w := \mathcal{N}(0,\lambda^{-1}I)\) \Comment{Initialize prior}
    \For{each \((R_i, F^{\text{real}}_{\text{peak},i}) \in\) TrainSet}
        \State \(L_i := p(F^{\text{real}}_{\text{peak},i} \mid R_i, w)\)
        \Comment{Likelihood estimation}
        \State \(w := \dfrac{L_i \cdot w}{p(F^{\text{real}}_{\text{peak},i} \mid R_i)}\)\Comment{Bayes' rule}
    \EndFor
    \State \(w_{\text{final}} := w\)

    \textbf{2-) Prediction Phase}
    \For{each \((R_j, F^{\text{real}}_{\text{peak},j}) \in\) TestSet}
        \State \( \hat{F}_{\text{peak},j} := \mathbb{E}[F_{\text{peak},j} \mid R_j, p(w_{final})] \)
        \If {\(\hat{F}_{\text{peak},j} \le \delta\)}
            \State \(Decision_j \gets\) \texttt{Pushable}
        \Else
            \State \(Decision_j \gets\) \texttt{Not Pushable}
        \EndIf
    \EndFor
\EndProcedure
\end{algorithmic}
\end{algorithm}

\section{EXPERIMENTS}

Our experiments are designed to demonstrate the accuracy of visual preliminary prediction and Bayesian affordance learning sections on both simulation and real-world scenarios. We have designed our own diverse simulation environment and real-world testbed having different slopes and rocks. We utilize a quadruped robot Unitree Go2 EDU with a robotic arm Unitree D1 mounted on top of it. All the hyperparameters described in the Section \ref{sec:methodology} are listed in Table \ref{tab:hyperparameters}. The robot's physical capabilities 
 $\delta$ is a very large set of parameters, however, in the scope of this study, we only include 3D dimensions, quadruped's maximum hoof elevation, step length, interfoot distance, and maximum force that robotic arm can generate.

\begin{table}[h]
\caption{Hyperparameters}
\label{tab:hyperparameters}
\begin{center}
\begin{tabular}{|l||c|}
\hline
\textbf{Hyperparameter} & \textbf{Value} \\
\hline
Bounding box scale factor ($s$) & 1.5 \\
Nearest neighbors count ($k$) & 30 \\
Neighbor search radius ($r$) & 0.2 m \\
Cosine similarity threshold ($T_{cs}$) & 0.85 \\
DBSCAN neighborhood radius ($\epsilon$) & 0.5 m \\
DBSCAN minimum points ($k_{\text{dbscan}}$) & 5 \\
Visual likelihood high threshold ($T_{\text{high}}$) & 0.8 \\
Visual likelihood low threshold ($T_{\text{low}}$) & 0.3 \\
\hline
\end{tabular}
\end{center}
\end{table}

To carry out simulation experiments, we prepared a Gazebo environment consisting of a height map terrain generated by Perlin noise, various sized rocks, and a GO2 robot with a mounted D1 servo arm. To construct the terrain, we created four different Perlin noise patches. While two of them were added to the flat height map to generate hills, the other two were subtracted to create pits. Patch positions were randomly selected. Six differently sized rocks were designed with the parameters shown in Table \ref{tab:rock_parameters}. In each experiment, the rocks were randomly rotated to achieve diversified contact surfaces. The collision geometry of the D1 robotic arm has been simplified to convex shapes to enhance simulation efficiency. A force-torque sensor is attached to the robotic arm to capture feedback from the forces generated during pushing actions.

\begin{table}[h]
\caption{Rock Parameters}
\label{tab:rock_parameters}
\begin{center}
\begin{tabular}{|l||c|c|}
\hline
\textbf{Rock Name} & \textbf{Scaling Factor} & \textbf{Weight} \\
\hline
Boulder 0  & (1, 1, 1) & 37.75 kg \\
Boulder 1   & (0.6, 0.6, 0.6) & 8.85 kg \\
Boulder 2  & (0.5, 0.5, 0.5) & 5.12 kg \\
Boulder 3   & (0.3, 0.3, 0.4) & 1.20 kg \\
Boulder 4   & (0.25, 0.25, 0.4) & 0.78 kg \\
Boulder 5   & (0.2, 0.2, 0.4) & 0.5 kg \\
\hline
\end{tabular}
\end{center}
\end{table}

\begin{figure}[t]
    \centering
    \includegraphics[ width=0.45\textwidth]{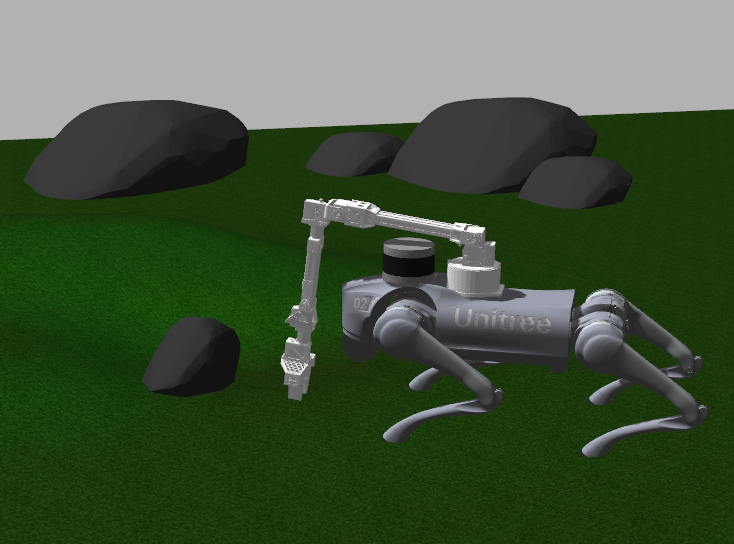}
    \caption{A rock-pushing experiment in the simulation environment is depicted. The rock is spawned on an uphill terrain in front of the robot. The robot moves closer to the rock and applies the predetermined pushing action.  }
    \label{fig:simulation}
\end{figure}

During each experiment as shown in Figure \ref{fig:simulation}, a rock is spawned in a random position and orientation. The robot starts from a pose facing the rock and is teleoperated by an expert toward the rock, where the robotic arm applies a predetermined push action. The point cloud of the scene is collected as a ROS bag during the robot's motion, while the force feedback data is gathered as a force signal during the push action. Each timestep in the point cloud ROS bag is added to the dataset along with the resulting force signal, enabling the inclusion of points captured at different times for the same setup. The features of the collected data set consist of obstacle position, box sizes, obstacle volume, obstacle shape score, and normal surface set. The maximum force-feedback magnitude for each run is recorded as the label for all data points in that run. We conducted 90 experiments, capturing both downhill and uphill pushes on the terrain generated by Perlin noise, resulting in a total of 857 data points.

\begin{figure}[h]
    \centering
    \includegraphics[ width=0.5\textwidth]{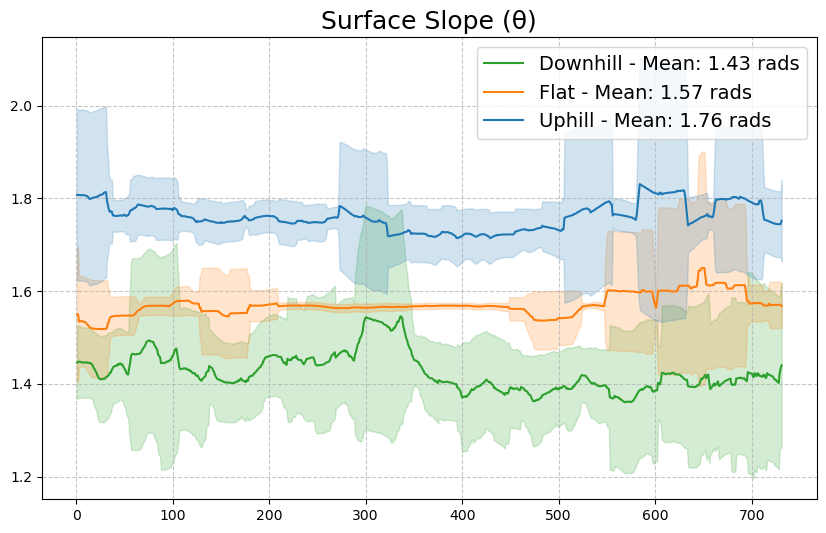}
    \caption{VPP distinguishes the slope of the surface in the simulation environment. Shady area represents the standard deviation.}
    \label{fig:surface_slope_estimation_sim}
\end{figure}
Preliminary visual prediction module both estimates volume of the obstacle and the surface normal with respect to robot's frame. In our experiments, VPP achieved to successfully differentiate the obstacles having higher volume from the smaller ones across various surface slopes. Similarly, VPP successfully estimates the average surface normal that obstacle is on as shown in the Figure \ref{fig:surface_slope_estimation_sim}. While it estimates angles around $\pi/2$ for the flat surfaces almost perfectly, the downhills and uphills produce a smaller and larger angles than the flat surface, respectively. These estimated values are fused using the Equation \ref{eqn:filtering} and the ones considered as pushable conveyed to the push affordance estimation module.

\begin{figure}[h]
    \centering
    \includegraphics[ width=0.46\textwidth]{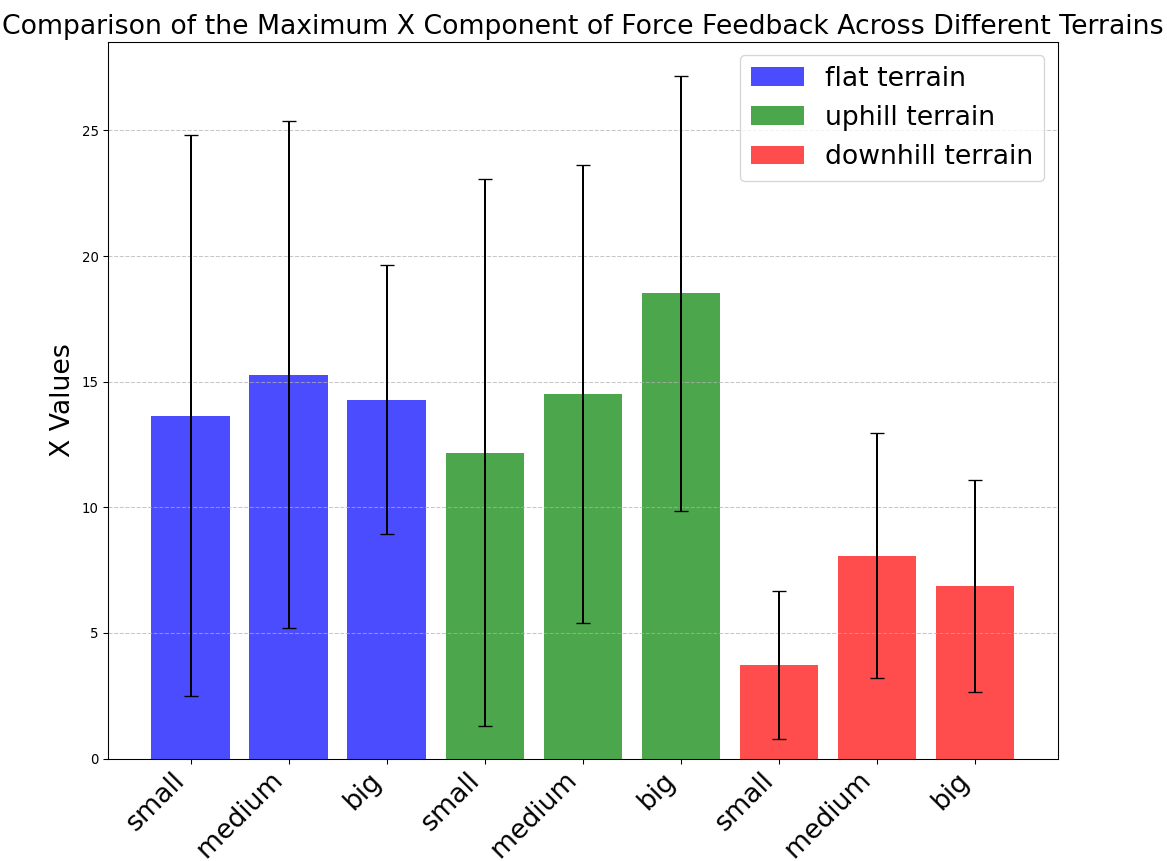}
    \caption{The mean and standard deviation of force feedback signals are presented for different combinations of rocks and terrain types. }
    \label{fig:data_analysis}
\end{figure}

The collected force feedback data is grouped according to the terrain on which the rocks rest and the rock size. The mean and standard deviation values are presented in Figure \ref{fig:data_analysis}. The high standard deviation is caused by the variety of contact points on the rock surface during the pushing action, as contact with different surfaces generates varying force feedback data in different directions. In the simulation experiments, it was empirically observed that if the magnitude of the sensed force feedback exceeds 20 N, the robot is unable to move the rock. This indicates that the pushability of rocks varies under different conditions and is highly dependent on surface properties.

\begin{figure}[h]
    \centering
    \includegraphics[ width=9cm]{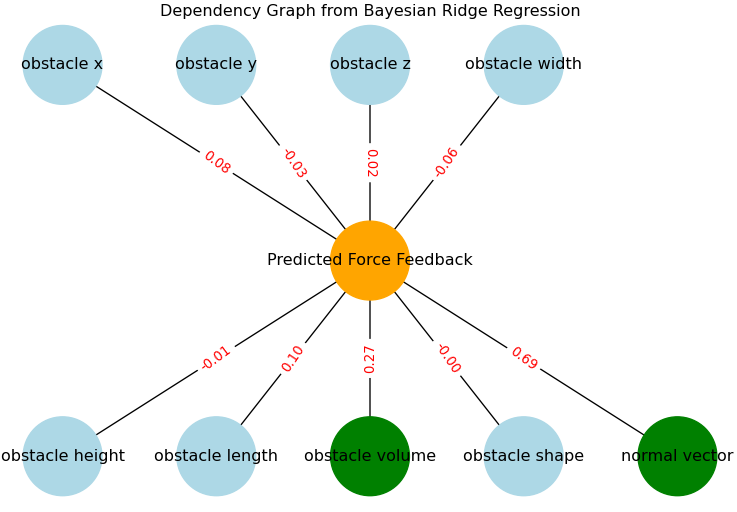}
    \caption{The dependency graph between the features and the predictions is presented. While the predictions are highly dependent on surface normals and obstacle volume, they are insensitive to the obstacle position at any observed time step. }
    \label{fig:result}
\end{figure}

The dataset is shuffled and split into training and test sets, with the test set comprising 30\% of the data. It is normalized using the StandardScaler function from the Scikit-learn library. The BayesianRidge model from Scikit-learn is used for fitting. The resulting coefficients and dependency graph are shown in Figure \ref{fig:result}. Given the provided features, the model predicts the maximum magnitude of the force feedback that will be sensed in a given configuration, which exhibits an inverse relationship with the pushability affordance of the rock. If the sensed force feedback value approaches the robot's limits, it strongly indicates that the rock is likely not pushable. Figure \ref{fig:result} strongly indicates that obstacle volume and the normal vector are proportional to the predicted force feedback. As the obstacle size and slope increase, the required force feedback also increases, reducing the rock's pushability. It is also observed that the spatial position of the rock in the point cloud is insignificant for prediction, meaning the robot can assess the rock's pushability at any timestep.

\begin{figure}[h]
    \centering
    \includegraphics[ width=0.5\textwidth]{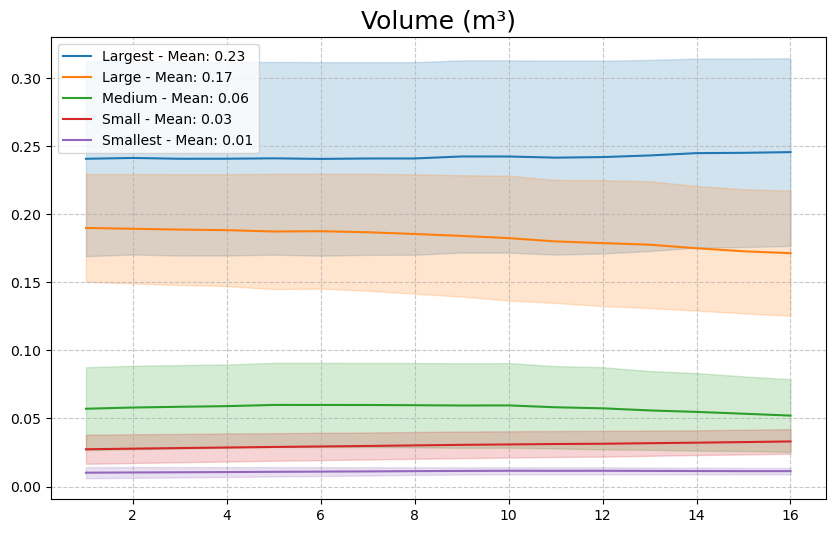}
    \caption{VPP accurately distinguishes the obstacles by their volume in the real world experiments. Shady area represents the standard deviation.}
    \label{fig:volume_estimation_real}
\end{figure}

To extend our work to the real-world applications, we collected point cloud from the stereo camera mounted on top of Unitree Go2 quadruped robot in our lab environment. We utilized five different artificial rocks with various sizes. Figure \ref{fig:volume_estimation_real} shows the estimated volumes of the rocks. Our VPP accurately finds the volume values from largest to smallest. We also utilized an additional surface having 5 degrees of slope. For the surface normal estimation, VPP finds the average surface normal of the uphill deviates from flat, which is perfectly estimated as $\pi/2$, as $0.1\pi$ which is roughly equal to 5 degrees, while our test ramp has 8 degrees of slope (Fig. \ref{fig:surface_normals}). Both of these graphs demonstrate the accuracy and the precision of our VPP in terms of both volume and surface normal estimation in real world data as well.

\section{CONCLUSION}

In this study, we have proposed a novel framework that determines the pushability of obstacles around a mobile agent. The objective of this framework is to facilitate the relocation of obstacles, as opposed to their avoidance, with the aim of enhancing long-term efficiency on routes that will be utilized repeatedly by multiple agents. The experimental phase of our study yielded encouraging statistical outcomes, underscoring the efficacy and precision of our system, as evidenced in both simulated and real-world settings. Our experiments demonstrate the efficiency of the designed visual estimation framework for pushability analysis and the correlation between required push action and obstacle/surface characteristics, such as volume and surface normal.

\subsection{Assumptions and Limitations}

In our study, we assume that when the robot pushes an obstacle, there is no slipping between the end effector and the obstacle surface. To achieve a robust pushing action, we can leverage multi-arm or multi-agent pushing to obtain force sensor feedback from various contact points. This approach allows us to better balance the trajectory of the rock being pushed and helps prevent slipping. Another assumption is that we consider the rocks to be detached from the ground and they are not outcrops. By further utilizing RGBD segmentation methods from the literature, we can enhance our study to distinguish between ground rocks and others.

\subsection{Lessons Learned}

The rocks in real-world environments have rough surfaces that consist of both concave and convex shapes, which differ from the widely studied everyday objects. Contact with these surfaces can result in translational movement, rotational movement, slipping of the end effector, or no movement at all, depending on the contact pose, orientation, and the local rock surface shape, as well as its normal vector. To further analyze the pushability of an obstacle, it is essential to provide detailed features of the local contact surface at the contact point. The movement of the rock is highly dependent on its center of mass and the sinkage ratio into the ground, which may induce rolling motion. Predicting the rock's center of mass and analyzing the terramechanics properties of the deformable terrain can further enhance our understanding of rock pushability.

In this study, we utilized a robotic arm for the pushing action. To handle heavier obstacles, we can take advantage of the high torque generated by the quadruped robot’s legs through brute-force actions such as jumping or kicking.

\subsection{Future Work}

To enhance our approach, we intend to integrate our visual preliminary prediction module with SLAM to obtain more comprehensive characteristics of the obstacles and the surface. Similarly, we plan to design an end-to-end deep neural network architecture for affordance learning to better represent complex relationships where linear models are insufficient. Subsequently, we intend to conduct high-level planning experiments in which the robot predicts the pushabilities of the rocks in the environment. This information will be used to enhance efficient trajectory planning and enable repeated navigation. Furthermore, our goals include to expand our real-world experiments to include a rough terrain with continuously and significantly varying slopes. Finally, incorporating the obstacle and surface material into the input set will enhance the usability of the framework on diverse surfaces, as the object-surface interaction is significantly influenced by friction.

\addtolength{\textheight}{-8cm}   





\end{document}